# Tenacious tagging of images via Mellin monomials

Kieran G. Larkin,[1,2,*] Peter A. Fletcher,[1] Stephen J. Hardy,[3]

[1]*Canon Information Systems Research Australia, Pty., Ltd, 1 Thomas Holt Drive, North Ryde, NSW 2113, Australia*
[2]*Nontrivialzeros Research, 22 Mitchell Street, Putney, NSW 2112, Australia*
[3]*National Information and Communications Technology Australia, ATP Level 5, 13 Garden Street, Eveleigh NSW 2015, Australia*
*\*Corresponding author: [iresearch@nontrivialzeros.net.](iresearch@nontrivialzeros.net.)*

We describe a method for attaching persistent metadata to an image. The method can be interpreted as a template-based blind watermarking scheme, robust to common editing operations, namely: cropping, rotation, scaling, stretching, shearing, compression, printing, scanning, noise, and color removal. Robustness is achieved through the reciprocity of the embedding and detection invariants. The embedded patterns are real one-dimensional Mellin monomial patterns distributed over two-dimensions. The embedded patterns are scale invariant and can be directly embedded in an image by simple pixel addition. Detection achieves rotation and general affine invariance by signal projection using implicit Radon transformation. Embedded signals contract to one-dimension in the two-dimensional Fourier polar domain. The real signals are detected by correlation with complex Mellin monomial templates. Using a unique template of 4 chirp patterns we detect the affine signature with exquisite sensitivity and moderate security. The practical implementation achieves efficiencies through fast Fourier transform (FFT) correspondences such as the projection-slice theorem, the FFT correlation relation, and fast resampling via the chirp-z transform. The overall method utilizes orthodox spread spectrum patterns for the payload and performs well in terms of the classic robustness-capacity-visibility performance triangle. Tags are entirely imperceptible with a mean SSIM greater than 0.988 in all cases tested. Watermarked images survive almost all Stirmark attacks. The method is ideal for attaching metadata robustly to both digital and analogue images. © 2013 Optical Society of America
  http://dx.doi.org/

## 1.Introduction

WATERMARKING (WM) of images was a popular subject of research in the 1990s and early 2000s. However, it was soon realized that each new method promoted a new attack, then enhanced methods, then counterattacks and so on in an endless arms race. Truly secure and robust WM is no longer considered possible by many experts, although agreement is not universal [1, 2] It is generally agreed that true security is achieved via cryptographic strength, and not through obscurity (this is known as Kerckoff's principle; see for example p291 Cox book [3]). In this paper we are not concerned with strong security against malicious attacks, but rather with data embedding that can survive all likely image degradations met in the everyday (specifically non-malicious) manipulation of images. An important use of such data embedding is the robust tagging of images with metadata (page 11, Cox book [3]). Metadata is so easily lost in many innocent image manipulations that a stronger binding to the image is sought by users. However in all cases it is essential that the tagged image show no perceptible deterioration relative to the original.

There has been a considerable literature devoted to watermarking methods that resist geometric distortion. Surprisingly the majority of published WM methods fail to survive image scaling, rotation, and (especially) cropping. The textbook by Cox et al [3] contains a summary of geometric distortion (page 269) related issues in WM as well as attacks based on distortion (page 303), also known as synchronization attacks.

Perhaps the archetypal digital image watermark is the spread-spectrum mark of Tirkel [4]. Spread-spectrum signals can be embedded at extremely low levels because the matched filter detection techniques boost signal levels by factors equal to the number of pixels (and noise by the square roots of the number of pixels). Detectable signals can be hundreds of times smaller than the image pixel values. In 1949 Shannon [5] observed that waveforms which fully utilize system (time-bandwidth) capacity "would be in all respects similar to white Gaussian noise". A little known patent of Pugsley [6] anticipates image watermarking with a method for hiding low amplitude signals in a printed image for the purpose of registration in a web printing press.

A common misconception (that we will try to dispel) in the watermarking literature is the spurious notion is that some watermarks can only be embedded in a transform domain (e.g. DCT, wavelet, Fourier, or Fourier Mellin). Johnson and Katzenbeisser [7] state: "… embedding information in the frequency domain of a signal can be much more robust than embedding … in the time domain." This has often been misconstrued to mean that the embedded signal exists in one domain and not the other. In reality a frequency domain watermark has a well-defined spatial (or time) domain equivalent that can be embedded additively in the spatial (or time) domain. The best known Rotation, Scale, and Translation (RST) invariant watermarking techniques [8, 9] only describe embedding in the transform domain. It can be shown that (multiplicative) Fourier domain embedding is equivalent to spatial convolution of Mellin basis functions with a phase-only image. Appendices A and B outline the previously unpublished, non-transform domain equivalents of the well-known RST invariant methods of O'Ruanaidh [9] and Lin [8] respectively.

Our method is an example of a blind method, i.e. the watermark is detectible without access to the original image. Our work is inspired by earlier methods which embed imperceptible templates, such as the Fourier point constellation embedding of Deguillaume et al [10] which uses a Hough transform for affine resistant detection. Similarly the method of Fleet et al [11] embeds two or more 2-D functions each with 1-D sinusoidal variation, and a Fourier transform with two or more spectral peaks; easily detected by an FFT. Perhaps the closest technique is that of Pereira and Pun [12] where the Fourier template comprises fourteen spectral points spread over two radial lines.

Often watermarking research concentrates on designing security then trying to add robustness afterwards. More recently Xie et al [13] start from robustness then modify the method to enhance security. This is very much our approach, but with the emphasis very much on robustness.

### 1.1 Contribution and the Structure of this paper

We would like to be absolutely clear about the motivation for, and significance of this paper. We propose a spatially direct way to add persistent tags to photographic images. Our motivation is to enable watermark technology for universal image labeling. As far as we are aware this is the first research to propose the use of 1-D Mellin monomials (viz. corrugated hyperbolic chirps) for watermarking. Our tagging technique combines a number of unfamiliar and seemingly unrelated mathematical elements seldom seen in the mainstream watermarking literature, so we have organized the description into the following sections:

Section 2 outlines the requirements of a robust image tagging scheme.
Section 3 presents the mathematics of spread spectrum Mellin monomials.
Section 4 relates the projection-slice theorem and correlation detection
Section 5 introduces the enhanced detectability of pattern constellations
Section 6 outlines the complete embedding and detection process.
Section 7 reports on the attack resistance and image quality of the new method.
Section 8 concludes.
Appendices A1 an A2 present, for the first time, spatial domain interpretations of two classic transform domain watermarking methods.

## 2. Requirements for Robust Tagging

### 2.1 Problems

In a previous work [14] we described a system for the direct embedding of basis patterns which combine spread-spectrum, and spread-space properties with rotation and scale invariance. Detection by matched filtering (namely correlation) then introduces the translation T into RST invariance. Unfortunately both printing and scanning operations typically introduce an aspect ratio change of several per cent and this drastically degrades the spread spectrum correlation peak height in practice. For example the LRHF correlation peaks fall by 55% following a 1% anamorphic magnification change. Even worse, the Pseudo Random Noise (PRN) correlation peaks fall by 80% following a 1% anamorphic magnification change (assuming a 512x512 image with a flat spectrum out to Nyquist frequency). Unless the WM system intrinsically compensates for such resizing the WM signal strength must be increased by a factor that fully compensates the correlation peak drop. But this inevitably leads to highly visible watermarks. Accordingly printing and scanning resistance demands a method immune to more general affine distortion.

### 2.2 Desirable properties

Ideally we would like to maintain all the advantages of our previous RST invariant watermarking system, but somehow add two more degrees of invariance, related to the affine spatial distortion. The following are the desirable system specifications:

A. Direct (additive) embedding of signal in image spatial domain (for speed);
B. Blind detection ("public watermarking system");
C. Direct detection for authorized users (not exhaustive search through all parameters);
D. Spread-spectrum and spread-space signal (to avoid trivial detection and attacks);
E. Invariance to affine spatial distortions over a wide range of parameters;
F. Insensitive to additive noise;
G. A large space of parameters for unauthorized users to search;
H. Insensitive to severe cropping and all the common graphical image modifications;
I. Capacity to embed useful metadata (e.g. 64 bits) at imperceptible signal levels;

Item E is the only change from our previous RST invariant system [6]. It is not immediately obvious how the system can be changed to accommodate the new requirement. Regrettably the logarithmic radial harmonic functions from the previous method are simply not invariant to general affine distortion.

## 2.3 Design Methodology

The watermarking literature contains numerous ingenious methods for achieving invariance. None to our knowledge satisfy all the above requirements. Affine invariance is the essential new feature. Consider the question; can the additional degrees of invariance be spread between the embedding and detection operations? Surprisingly the answer is yes.

Firstly we consider embedding. A number of different spatial patterns have been proposed for RST invariant watermarks [15]. Under a general affine distortion a pattern transformation is fully defined by 6 parameters. A 1-D function can be extended over the plane (or to use Bracewell's more evocative term *corrugated* [16]) so that it only undergoes a 3 parameter transformation under affine distortion. If the 1-D function is itself scale invariant, then just 2 parameters fully define the transformation: orientation and distance-from-the-origin of the pattern's center-line. Hence a scale invariant corrugated pattern can be found through just two dimensions of search after the application of an affine distortion. Furthermore, if the scale invariant function is spread spectrum, then there is the possibility of a correlation based search strategy.

Secondly we consider detection. The Radon transformation is an invertible image projection, converting Cartesian to polar coordinates, where polar coordinates define orientation and distance from origin. Under Radon transformation a 2-D corrugated scale invariant pattern collapses down to a 1-D lineal projection. The energy in a 1-D projection can be further concentrated down to a point by 1-D correlation, if the function is spread spectrum.

The preceding embedding and the detection operations require a substantial amount of mathematical theory unfamiliar to the watermarking literature, which we outline in the next three sections.

## 3. Properties of 1-D Mellin Monomials

In this section we show that certain complex monomials have appropriate invariance.

### 3.1 Hyperbolic Chirps, Mellin Monomials, Homogeneous Functions, and Scale Invariance

First we need to clarify our terminology. The functions we propose have many names, depending on application. Our previous work [14] used the naming popular in optical target recognition; the Logarithmic Radial Harmonic Function (LRH), introduced by Rosen et al [17] after the influential work of Casasent & Psaltis [18]. The classic work of Gelfand and Shilov [19] describes the Fourier properties of monomials (or complex-index homogeneous functions) which, by definition, exhibit dilation invariance. It is no coincidence that these 1-D functions are eigen-functions of Baraniuk's dilation operator [20], eigen-functions of the quantum correlation operator of de la Torre [21], and orthogonal functions defining the Mellin transform [22]. Inspired by the Doppler tolerant sonar of certain bat species Altes notes that hyperbolic chirp functions (HCFs) have scale invariant properties related to the Fourier-Mellin transform (FMT) [23]. For expediency we have chosen to use the shortest name, Mellin monomial (MM), to denote the basis functions in this work.

The main properties of the 2-D scale invariant LRHFs have been derived in our previous WM work [14]. Champeney [24] covers the Fourier properties (existence, convergence, Hilbert spaces etc.) of the 1-D Mellin monomials. Firstly note that there are two quite independent and orthogonal forms, one with even parity and one with odd, respectively:

$$h_s^+(x) = \frac{|x|^s}{|x|^{1/2}} \equiv \frac{\exp(s \log|x|)}{|x|^{1/2}}, \qquad h_s^-(x) = \text{sgn}(x) \cdot \frac{|x|^s}{|x|^{1/2}} \equiv \frac{\text{sgn}(x)}{|x|^{1/2}} \cdot \exp(s \log|x|) \qquad (1)$$

It is well known that monomials with real powers are homogeneous. Perhaps less known is when the monomial index $s = \lambda + 1/2 + i\sigma$ is extended to the complex domain the function remains homogeneous (although the real and imaginary parts now oscillate wildly). If a function is scaled by a real positive factor, then it follows that:

$$h_s^-(ax) = c.h_s^-(x), \qquad h_s^+(ax) = c.h_s^+(x), \qquad c = a^{s-1/2} = e^{(s-1/2)\log a} \qquad (2)$$

Hence the functions are homogeneous of degree $\{s - 1/2\}$. Perhaps the most surprising property is the Fourier transform. The usual Fourier inverse scaling property is seemingly subverted by the homogeneous property (as we will explain later on). Champeney gives the following definition of the 1-D Fourier transform:

$$F(u) = \int_{-\infty}^{+\infty} f(x) \exp(-2\pi i u x) dx, \quad F(u) \xleftrightarrow{FT} f(x), \quad H_s^{\pm}(u) \xleftrightarrow{FT} h_s^{\pm}(x) \qquad (3)$$

It then follows that the Fourier transforms we seek exist in the space of tempered distributions $D'$, which we write in the symmetric index forms:

$$\frac{|x|^s}{|x|^{1/2}} \xleftrightarrow{FT} K^+ \frac{|u|^{-s}}{|u|^{1/2}}, \qquad \text{sgn}(x) \frac{|x|^s}{|x|^{1/2}} \xleftrightarrow{FT} K^- \text{sgn}(u) \frac{|u|^{-s}}{|u|^{1/2}} \qquad (4)$$

The constants being given by:

$$K^+ = +2 \frac{\Gamma(s - 1/2)}{(2\pi)^{(s-1/2)}} \cos\left(\frac{\pi[s - 1/2]}{2}\right), \quad K^- = -2 \frac{\Gamma(s - 1/2)}{(2\pi)^{(s-1/2)}} \sin\left(\frac{\pi[s - 1/2]}{2}\right) \qquad (5)$$

The Fourier relations break down for integer values of $\lambda = \Re(s-1/2)$, and three special cases occur (interested readers should consult the tabulated Fourier transforms in chapter 13 of Champeney [24]).

### 3.2   Mellin Monomials and Orthogonality

One of the many fascinating properties of Mellin monomials is their approach toward infinite phase gradient near the origin. As these signals will be embedded in digital images there will be a central region that cannot be reproduced without aliasing, so it is set to zero in our method. Fig.1 shows an MM with the characteristic hyperbolic frequency:

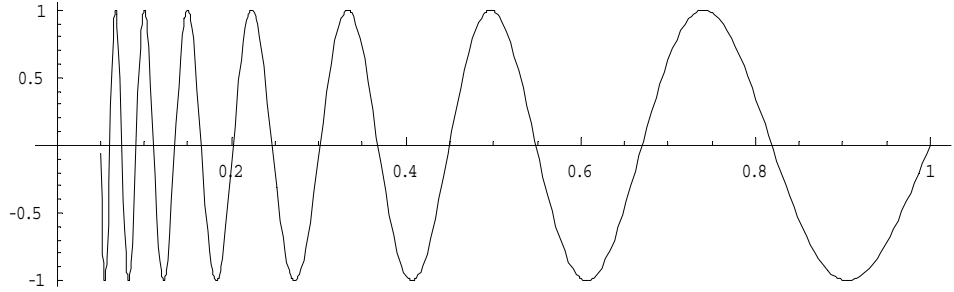

Fig.1. One side of a Mellin monomial function in the range $x = \varepsilon \to 1$

The orthogonality of MMs is an important issue, as is the question of how many orthogonal MMs can exist in a given system. Consider a complex, two-sided MM and its orthogonality to another MM represented by overlap integral $\Omega$:

$$\Omega(s_1, s_2^*) = \int_{-b}^{b} h_{s_1}^{\pm}(x) h_{s_2^*}^{\pm}(x) dx = 2 \int_{a}^{b} \frac{\exp\left(\left[s_1 + s_2^*\right]\log|x|\right)}{|x|} dx. \qquad (6)$$

Even MMs are automatically orthogonal to odd MMs by symmetry. Letting $\log b = B$, $\log a = -B$, then:

$$\Omega(s_1, s_2^*) = 2\left\{\frac{\exp\left(\left[(1+2\lambda)+i(\sigma_1-\sigma_2)\right]B\right) - \exp\left(-\left[(1+2\lambda)+i(\sigma_1-\sigma_2)\right]B\right)}{(1+2\lambda)+i(\sigma_1-\sigma_2)}\right\} \qquad (7)$$

The overlap integral is zero IFF the real and imaginary parts of the numerator are zero:

$$\lambda = -1/2, \qquad (\sigma_1-\sigma_2)B = m\pi \qquad (8)$$

The first condition $\lambda = -1/2$ is the Moses and Prosser [22] orthogonality condition. The second condition determines how many orthogonal functions exist for a finite signal domain $b = e^B, a = e^{-B}$. The highest non-aliased frequency occurs at the lower bound, and the phase $\psi(x) = \sigma \log|x|$ implies the phase derivative (i.e. frequency) is hyperbolic $d\psi/dx = \sigma/x$ and no greater than the Nyquist limit, that is $(\sigma/x) \le (\pi/\Delta)$. The number of orthogonal sine components in this range is equal to the maximum number of periods fitting in, namely $\sigma_{\max}(\ln b - \ln a)/2\pi = (\ln b - \ln a)/(2a)$. Thus a $1024^2$ pixel image with a 188 pixel lower bound (a=188$\Delta$, b=512$\Delta$) supports $188\ln(e)/2 = 94$ different MMs. Including even and odd symmetry allows twice as many modes. Keeping the least visible of these modes results in about 94 useable orthogonal modes overall. Assuming $\lambda = -1/2$, then:

$$\Omega(s_1, s_2^*) = 4B \operatorname{sinc}\left[(\sigma_1-\sigma_2)B\right] \qquad (9)$$

In the limit of MMs with infinite support ($B \to \infty$), the overlap integral approaches a Dirac delta distribution.

### 3.3   Mellin Monomials and Good Correlation

Overlap orthogonality is not equivalent to correlation orthogonality. Ideal correlation detection requires Dirac delta-like correlation for the chosen basis function (maximum energy concentration) and flat-line for all other basis functions (maximum energy dispersion). In practice the Welch lower bound on cross-correlation maxima applies [25]. In the Fourier domain a delta correlation has a flat spectrum and a constant phase. In contrast the Fourier domain of a flat-line can have a flat spectrum only if the phase fluctuates wildly. For illustration we consider an even MM with infinite support. The Fourier correlation theorem defines the Fourier spectrum and phase as follows:

$$c(x) \equiv \int_{-\infty}^{+\infty} \frac{|x'-x/2|^{s_1}}{|x'-x/2|^{1/2}} \cdot \frac{|x'+x/2|^{s_2^*}}{|x'+x/2|^{1/2}} dx' = \frac{|x|^{s_1}}{|x|^{1/2}} \otimes \frac{|x|^{s_2}}{|x|^{1/2}}$$

$$(10)$$

More conveniently in the Fourier domain:

$$c(x) \xleftrightarrow{FT} K^+ \frac{|u|^{-s}}{|u|^{1/2}} \left( K^+ \frac{|u|^{-s}}{|u|^{1/2}} \right)^* = |K^+|^2 \frac{|u|^{i(\sigma_1-\sigma_2)}}{|u|^{2(1+\lambda)}} \quad (11)$$

But we know (see [24] or [19]) that Fourier transformation just inverts and offsets the index of a homogeneous function:

$$c(x) = \kappa \frac{|x|^{-i(\sigma_1-\sigma_2)}}{|x|^{-1-2\lambda}}, \quad 2\lambda \notin \mathbb{Z}, \quad (12)$$

where $\kappa$ is a constant (related to $\lambda$). We have to be quite careful as there are a number of special cases for integer and half-integer real powers (but only when the imaginary part is zero). The most important case being the delta function correlation (or flat spectrum) condition when $\lambda = -1$.

$$c(x) = \begin{cases} \gamma \delta(x), & (\sigma_1-\sigma_2) = 0 \\ \kappa \frac{|x|^{-i(\sigma_1-\sigma_2)}}{|x|}, & (\sigma_1-\sigma_2) \neq 0 \end{cases} \quad (13)$$

Here $\gamma$ is a constant necessary to conserve signal energy. For the case of the overlap orthogonality condition, $\lambda = -1/2$, we find a logarithmic peak or spatial flat-line

$$c(x) = \begin{cases} const - \gamma \log|x|, & (\sigma_1-\sigma_2) = 0 \\ \kappa |x|^{-i(\sigma_1-\sigma_2)}, & (\sigma_1-\sigma_2) \neq 0 \end{cases}, \quad (14)$$

For the (highly practical) flat spatial envelope condition, $\lambda = 0$, we obtain a result symbolized by the inverse Laplacian, $(-\Delta)^{-1}$ or fractional integration of the Dirac delta [26, 27]:

$$c(x) = \begin{cases} \gamma (-\Delta)^{-1} \delta(x), & (\sigma_1-\sigma_2) = 0 \\ \kappa |x| |x|^{-i(\sigma_1-\sigma_2)}, & (\sigma_1-\sigma_2) \neq 0, \end{cases} \quad (15)$$

where $\gamma$ is a constant necessary to conserve total signal energy. The result can also be expressed in terms of Green's functions [19].
In all four cases above the matched filter condition $(\sigma_1-\sigma_2) = 0$ results in a highly peaked correlation function, whilst the non-match condition $(\sigma_1-\sigma_2) \neq 0$ results in a highly oscillatory and dispersed function with zero mean at the center. A finite domain size has a significant effect on the correlation; decohering the otherwise singular behavior near integer real indices. In practice we find that the real index has little effect on the orthogonality in the range $-1 \leq \lambda \leq 0$ and matched correlations are always sharply peaked functions. Phase correlation [28] and local phase enhancement [29] are two effective methods we use to whiten the Fourier spectrum and impose the preferred delta-like correlation peak.
Although true scale invariance generally requires a complex function, e.g. (3), it is possible to embed the real part of a MM as a watermark and detect it with a complex template MM and regain the invariance properties. The method was outlined in our previous RST WM paper [14]. Essentially the complex template correlates with one of the Hermitian pair of scale invariant patterns contained within a real MM:

$$\Re\left[h_s^+(x)\right] = \frac{|x|^s}{2|x|^{1/2}} + \frac{|x|^{s^*}}{2|x|^{1/2}} \quad (16)$$

The other Hermitian component is totally decorrelated and its signal is spread widely, with its frequencies doubled. Typically we embed within the luminance signal, which has the advantage that it survives color removal and major hue shifts and typically represents the underlying image structure well, and is often compressed using a higher bandwidth than the color signal. The watermark signal $w(\mathbf{r})$ comprises a collection of these corrugated scale invariant functions, with varying angles, offsets and phases, with a predetermined value of the imaginary index $\Im(s) = \sigma$, and symmetry of the basis (odd or even) is represented by $\chi$:

$$w(\mathbf{r}) = \sum_{k=1}^{K} \Re\left[h_s^\chi(\xi_n)\right]. \quad (17)$$

### 3.4 Corrugated Mellin Monomials and Partial Affine Invariance
Consider the action of a general affine transform on a corrugated function $g(\mathbf{r}) = f(\mathbf{r}.\hat{\mathbf{p}} - p)$.

The spatial Cartesian coordinates of our system are $(x, y)$ and the corresponding position vector is $\mathbf{r} = x\mathbf{i} + y\mathbf{j}$, and corrugation normal unit vector $\hat{\mathbf{p}} = \cos\beta\mathbf{i} + \sin\beta\mathbf{j}$ points at polar angle $\beta$. The corrugated 2-D function $g(\mathbf{r})$ can be generated by back-projection of a 1-D function $f(x)$ that is offset from the origin and rotated to align with the line defined by the vector equation $\mathbf{r}.\hat{\mathbf{p}} = p$. Consider the most general affine coordinate transform $\mathbf{r} \to \mathbf{r}'$ defined by six parameters:

$$\begin{pmatrix} x' \\ y' \end{pmatrix} = \begin{pmatrix} a_{11} & a_{12} \\ a_{21} & a_{22} \end{pmatrix} \begin{pmatrix} x \\ y \end{pmatrix} + \begin{pmatrix} x_0 \\ y_0 \end{pmatrix} = \mathbf{a} \begin{pmatrix} x \\ y \end{pmatrix} + \begin{pmatrix} x_0 \\ y_0 \end{pmatrix} \quad (18)$$

The inverse affine transformation is similarly written. The corrugated function $f(\mathbf{r}.\mathbf{p} - p) = f(\xi)$, has a similar form after affine distortion $f(\xi) = f(\mathbf{r}'.\mathbf{p}' - \tilde{p})$ mapping to a rotated, scaled, and shifted function. The three changed parameters are reduced to just two if the corrugated function is scale invariant. Thus an affine transformation with six independent transform parameters *only* induces rotation and offset in a scale invariant corrugated function.

## 4 The Radon Transform and Projection-Slice Theorem

Our previous work on RST invariant watermarks utilized 2-D cross-correlation for direct detection. Correlation has the great advantage of essentially collecting *all* the energy of an embedded pattern and concentrating it into a single correlation peak. Applying this technique to corrugated scale-invariant functions gives a 2-D correlation applied over a 2-D search space, resulting in 4-D complexity overall. Lower complexity is desired, and the Radon transform affords a solution by projecting each 2-D basis onto 1-D.

### 4.1 Fourier Transform of a Corrugated Mellin Monomial

The Radon transform may be applied *in-place* (viz. in the same Cartesian space as the image) or, as is more conventional, in the polar coordinates. In the following analysis we take advantage of the projection-slice theorem [30]. The theorem states that (in 2-D) the *projection* of a function onto the perpendicular radial line is the Fourier transform of the corresponding radial *slice* in the Fourier transform domain. Readers interested in more details should consult the exposition of Strichartz [31] or the Bracewell textbook [16]. Corrugated basis functions allow us to elude the usual complications of discrete (Radon) projection because the Fourier transform of the corrugated MM is simply a radial slice, as we now show.

The 2-D Fourier transform is defined in Cartesian or vector coordinates as follows:

$$G(u,v) = G(\mathbf{q}) = \int_{-\infty}^{+\infty}\int_{-\infty}^{+\infty} g(x,y) \exp\left[-2\pi i(ux+vy)\right] dx dy = \int_R g(\mathbf{r})\exp\left[-2\pi i \mathbf{r}.\mathbf{q}\right] d\mathbf{r} \quad (19)$$

We also denote the 2-D Fourier transform $G(u,v) \xleftrightarrow{2-D\ FT} g(x,y)$. Equations (3) and (4) show the 1-D FT of a basis function $H_s^{\pm}(u) \xleftrightarrow{1-D\ FT} h_s^{\pm}(x)$, hence the 2-D FT can be written:

$$G(\mathbf{q}) = \int_R h_s^{\pm}(\mathbf{r}.\hat{\mathbf{p}}_n - p_n)\exp\left[-2\pi i \mathbf{r}.\mathbf{q}\right] d\mathbf{r} \quad (20)$$

Define spatial and Fourier coordinates rotated by angle $\beta_n$:

$$\begin{cases} \mathbf{q} = u\mathbf{i} + v\mathbf{j} = q\cos\varphi\mathbf{i} + q\sin\varphi\mathbf{j} \\ \mathbf{q}' = u'\mathbf{i} + v'\mathbf{j} = q\cos(\varphi-\beta)\mathbf{i} + q\sin(\varphi-\beta)\mathbf{j} \end{cases} \quad \begin{cases} \mathbf{r} = x\mathbf{i} + y\mathbf{j} = r\cos\theta\mathbf{i} + q\sin\theta\mathbf{j} \\ \mathbf{r}' = x'\mathbf{i} + y'\mathbf{j} = r\cos(\theta-\beta)\mathbf{i} + r\sin(\theta-\beta)\mathbf{j} \end{cases} \quad (21)$$

Rotating FT (22) to align with the corrugation normal $\hat{\mathbf{p}}_n$ using relations (22) gives

$$G(\mathbf{q}') = \delta(v')\int_{-\infty}^{+\infty} h_s^{\pm}(x'-p_n)\exp\left[-2\pi i u'x'\right]dx' = \exp\left[-2\pi i u'p_n\right]\delta(v')H_s^{\pm}(u') \quad (22)$$

Essentially the Fourier transformation has collapsed the 2-D corrugate function into an axial slice with a linear phase modulation proportional to the offset $p_n$. Fig. 2 shows the resulting Fourier slice, perpendicular to the spatial corrugations.

The slice direction $u'$ is parallel to the corrugation normal $\hat{\mathbf{p}}_n$. In the next section we'll see some figures showing the result in (22). The analysis in this section is described in the continuous domain for brevity; in other words "think analog, act digital" [32]. Discrete implementation introduces artefacts familiar to users of the discrete Fourier transform (DFT). However, we have found the method to be remarkably resilient to such effects.

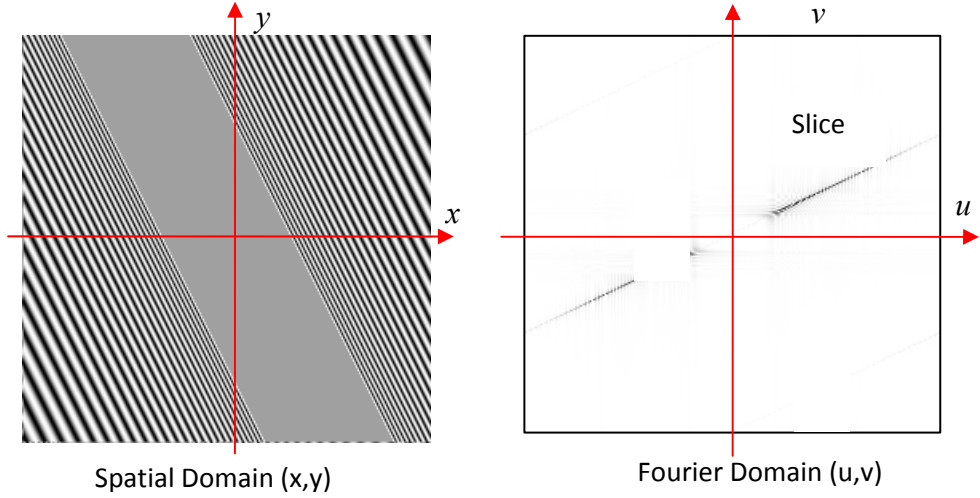

Fig.2    One basis function (left) and the Fourier transform magnitude (right)

### 4.2    Correlation Detection

Each radial Fourier slice corresponds to a spatially dispersed watermark basis pattern. Fig. 3 shows a typical slice arrangement. To detect each basis ideally we need a 1-D correlation with the projection of a basis pattern in the spatial domain. Taking advantage of the Fourier correlation theorem we see this is equivalent to conjugate multiplication in the Fourier domain. As we do not know the orientations *a priori* we must repeat the 1-D correlation over all possible orientations.

An efficient way to do this is via the polar transform of the Fourier transform, shown in Fig.4. Note that the bilateral polar transform allows positive and negative values of the Fourier radial coordinate q. This means that the orientation is only needed in the range $|\varphi| \leq \pi/2$. The bilateral polar transform is also known as the quasi-polar transform and it has the benefit of removing the complex Hilbert transform relationship due to the one-sidedness of the conventional polar radius $|q|$.

We can summarize the overall correlation detection process as follows. Firstly the input image is Fourier transformed:

$$g(x,y) \xrightarrow{\text{FT}} G(u,v) \quad (23)$$

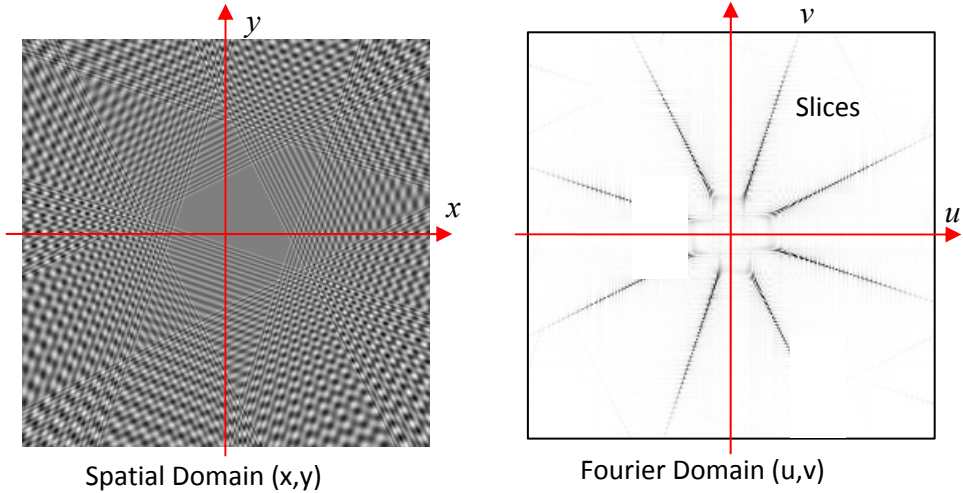

Fig.3 Four basis function (left) and their Fourier transform magnitude (right)

The quasi-polar coordinate transform is defined over all directions, although it is only needed over 180°: $\varphi = \text{atan2}(v,u)$, $|\varphi| \leq \pi/2$, $q = \text{sgn}(\cos\varphi)\sqrt{u^2+v^2}$. We then express the Fourier transform in quasi-polar coordinates

$$\mathsf{G}(q,\varphi) = G(u,v) \quad (24)$$

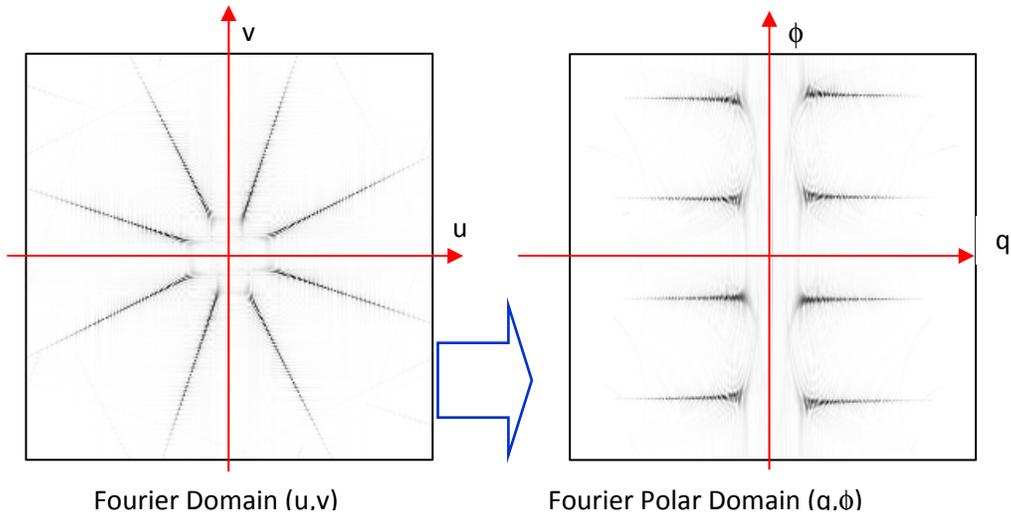

Fourier Domain (u,v)　　　　　　　　　Fourier Polar Domain (q,ϕ)

Fig.4 Four Fourier basis functions (left) and their polar transform (right)

Now 1-D correlation is repeated over all orientations $\phi$ by Fourier multiplication

$$G(q,\varphi).H^*(q) \xrightarrow{\text{1D-IFT q-p}} g(p,\varphi) \otimes h(p) \tag{25}$$

Explicitly writing the 1-D inverse gives the correlation over offset and orientation $(p,\varphi)$ :

$$C(p,\varphi) = \int_{-\infty}^{+\infty} G(q,\varphi).H^*(q)\exp[+2\pi ipq]dq \tag{26}$$

Fig.5 shows the result of such a correlation on the four basis patterns. Note the four strong correlation peaks defining the four center-lines in the correlation domain $(p_n,\varphi_n)$.

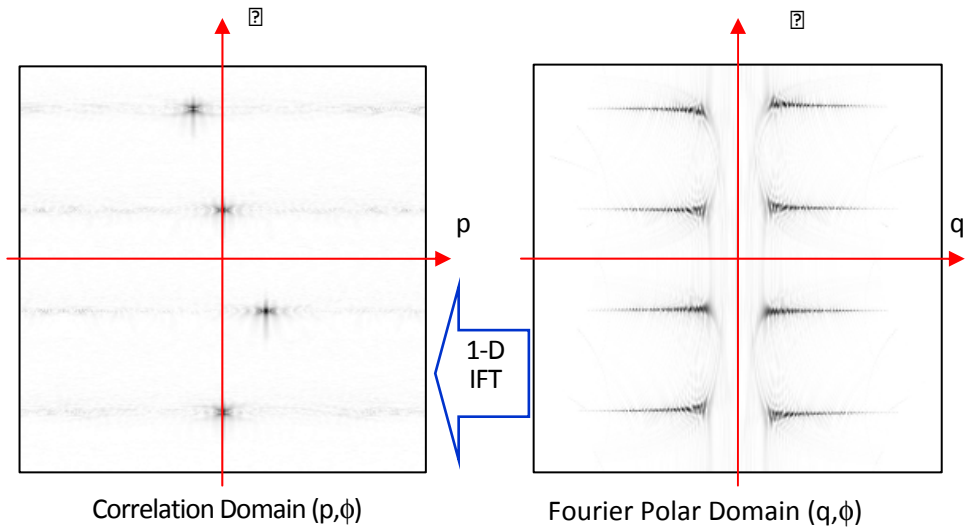

Correlation Domain (p,ϕ)　　　　　　　Fourier Polar Domain (q,ϕ)

Fig.5 Correlation in a polar spatial domain (offset p, direction ϕ) and Fourier polar equivalent

It is worth noting here that a common problem with the many watermarking techniques based on image moments is that they are inherently unable to survive even small amounts of image cropping. The problem stems from the sensitivity of moments to boundary conditions. Correlation, on the other hand, is much less sensitive to boundary conditions, and can easily be made even less sensitive by windowing.

### 4.3 Fast Pseudo-Polar Mapping

It transpires that the exact 2-D Cartesian to polar mapping is fraught with complications. We have developed a scheme, a similar to that described by Mersereau and Oppenheim [33]. More recently Averbuch et al [34] have shown the accuracy and speed advantages. Their pseudo-polar transform allows re-gridding with the perfect sinc interpolation properties of the chirp-z transform. The chirp-z facilitates fast (order NlogN) resampling with arbitrary rescaling factors, including irrational factors, on the nested squares of the pseudo-polar grid. One complication of the pseudo-polar transform is that the angular and radial sampling is slightly non-uniform, although radial sampling is uniform within each slice. This does not create any problems for scale invariant functions, and so we omit any second interpolation stage to create a uniform sampling.

## 5 Enhanced Detectability of Basis Pattern Constellations

To increase template detection sensitivity and discrimination it is desirable to embed constellations of MMs so that the configurations themselves are also affine invariant. If such configurations exist then we could conceivably reduce the visibility of the embedded tag for any predefined robustness level. An affine distortion is parameterized by six independent coefficients but a single MM encodes just two. Under an affine transformation the ratio of lengths of two collinear line segments is invariant. Four basis patterns define four lines with up to six line intersections and four pairs of line segments, hence four invariant ratios. We can use these ratios to uniquely and unambiguously determine any affine transformation. Consider Fig.6. The four affine invariant ratios are $\gamma_n = a_n/b_n$. If reflections are included as possible affine distortions, then the arrangement should be strongly chiral. This particular example covers the angular spectrum uniformly in 45° steps. Six intersection points give an over-determined affine transformation. But how do we initially identify the affine invariant ratios?

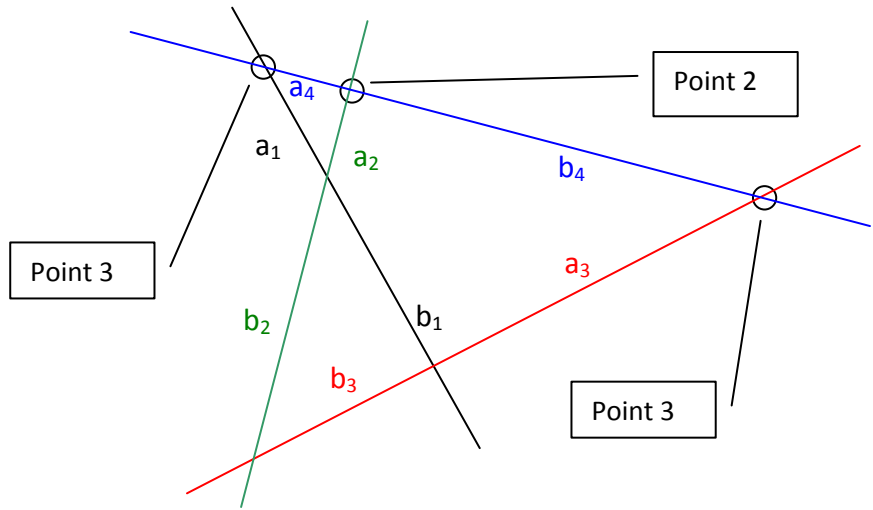

Fig.6 The center-lines of four MMs and their corresponding line segment ratios a/b.

The output from the previous correlation stage is a list of line equations defining the most likely center lines of embedded basis patterns. We take all possible quartets from M detected correlation peaks. For each of the four lines compute the three line intersections and thus the line segment ratio for each line. This gives four ratios. If any ratio is greater than unity it is inverted and we generate an ordered list of four ratios. This list is a signature or *fingerprint* of the template. It is possible to design the 45° template with four maximally spaced and unique ratios, thus allowing us to distinguish each line and intersection. Further details can be found in the patents of Larkin et al [35, 36]. Summarizing: each line has a unique ratio and each point is the unique intersection of two lines. The number of quartets chosen from M prospective points is $C_4^n = N!/((N-4)!4!)$. Thus 20 potential lines give less than 5000 ($C_4^{20} = 4845$) ordered lists to check for the ratio signature. This operation is not very time-consuming. Finding one or more ratio signatures close to the desired signature means we can move on to the next step – computing the best estimate of the affine transformation from the six intersection points. If two lines with normal angle $\beta$ and offset distance from origin $p$ are separately indexed by $k$ and $m$, then the intersection point $(x_{km}, y_{km})$ is given by the following equation:

$$\begin{pmatrix} x_n \\ y_n \end{pmatrix} = \begin{pmatrix} x_{km} \\ y_{km} \end{pmatrix} = \frac{1}{\sin(\beta_k - \beta_m)} \begin{pmatrix} \sin\beta_k & -\sin\beta_m \\ \cos\beta_k & \cos\beta_m \end{pmatrix} \begin{pmatrix} p_m \\ p_k \end{pmatrix}, \ k=1 \to 4, \ m \neq k, \ n=1 \to 6 \qquad (27)$$

Obviously intersecting lines are non-parallel. We chose the four lines to be maximally intersecting, which means equi-spaced in angle or approximately 45° apart.

## 5.1 Constellation verification

Assuming that one of the combinations of four points is the correct combination, how do we ascertain the most likely combination? The correlation strength is one indicator of likelihood (but is easily affected by lines and edges in an image). However, a combination that contains ratios that deviate significantly from the expected ratios is unlikely to be a valid combination. We can simply set an allowable range of values for the ratios and only consider combinations within that range. We take this approach with an allowable ratio deviation of $\pm 0.02$. An overall criterion is based on the smallest relative ratio Euclidean distance:

$$E_{ijkl}^2 = \sum_{n=i,j,k,l} (r'_n - r_n)^2 / r_n^2 \tag{28}$$

A further validity test is to make sure that the implied affine transformation is plausible.

## 5.2 Affine parameter estimation

From four plausible ratios we can infer an affine transformation. The four ratios uniquely assign to four corresponding lines. Equation (27) allows us to compute the 6 intersection points, taking two lines at a time. As long as no ratio is too close to 0.5 we can unambiguously identify points. We can now compute a least squares affine estimate. Note that reflection as a valid affine transformation. An estimate for the misalignment of actual points $(\tilde{x}_n, \tilde{y}_n)$ and affine transformed ideal points $(x'_n, y'_n)$ is given by:

$$T = \sum_{m=1}^{6} (\tilde{x}_n - x'_n)^2 + (\tilde{y}_n - y'_n)^2 \tag{29}$$

From (19) we have from the original ideal points $(x_n, y_n)$

$$\begin{pmatrix} \tilde{x}_n \\ \tilde{y}_n \end{pmatrix} = \begin{pmatrix} x'_n \\ y'_n \end{pmatrix} + \begin{pmatrix} \varepsilon_x \\ \varepsilon_y \end{pmatrix} = \begin{pmatrix} a_{11} & a_{12} \\ a_{21} & a_{22} \end{pmatrix} \begin{pmatrix} x_n \\ y_n \end{pmatrix} + \begin{pmatrix} x_0 \\ y_0 \end{pmatrix} + \begin{pmatrix} \varepsilon_x \\ \varepsilon_y \end{pmatrix}, \quad n = 1 \to N \tag{30}$$

Minimizing T with respect to the unknown affine parameters gives a sequence of equations:

$$\begin{pmatrix} \sum_{n=1}^{N} \tilde{x}_n x_n & \sum_{n=1}^{N} \tilde{y}_n x_n \\ \sum_{n=1}^{N} \tilde{x}_n y_n & \sum_{n=1}^{N} \tilde{y}_n y_n \\ \sum_{n=1}^{N} \tilde{x}_n & \sum_{n=1}^{N} \tilde{y}_n \end{pmatrix} = \begin{pmatrix} \sum_{k=1}^{N} x_n x_n & \sum_{k=1}^{N} x_n y_n & \sum_{k=1}^{N} x_n \\ \sum_{k=1}^{N} y_n x_n & \sum_{k=1}^{N} y_n y_n & \sum_{k=1}^{N} y_n \\ \sum_{k=1}^{N} x_n & \sum_{k=1}^{N} y_n & \sum_{k=1}^{N} 1 \end{pmatrix} \begin{pmatrix} a_{11} & a_{21} \\ a_{12} & a_{22} \\ \varepsilon_x & \varepsilon_y \end{pmatrix} = \mathbf{M} \begin{pmatrix} a_{11} & a_{21} \\ a_{12} & a_{22} \\ \varepsilon_x & \varepsilon_y \end{pmatrix} \tag{31}$$

Then we simply invert the 3x3 matrix to yield the six affine parameters. Incorrect labeling of the points will result in large position errors for some of the matches and hence large overall error term $T$. This makes it easy to filter unlikely configurations in the detection process. We also restrict the range of allowable affine transformations by restricting affine parameters $a_{jk}$.

# 6 Overview of the Tagging and Decoding Processes

The overall watermark process can be summarized in two figures. The first, Fig.7 shows the embedding process. The second, Fig.8 shows the detection process.

## 6.1 Embedding

A set of four MMs are generated using a key to set the imaginary index of the function. The four line-intersection ratios represent another simple key, and a full 2-D template image is generated.

The watermark message (or payload) can also be embedded using additional MM basis patterns, or indeed any convenient spread spectrum (SS) patterns. In our current implementation the actual message or metadata is embedded using conventional pseudo-random noise (PRN) patterns. Advantages are that the message is cryptographically secure, and all the available bandwidth is used. Disadvantages are that two separate systems of patterns are embedded, requiring two different decoding passes, and that it is also necessary to perform an affine transform to restore the image to the original embedding coordinate system, or one related to it, before correlation detection is possible.. If the template is attacked and removed, then the resistance of the metadata to distortion is compromised, and the message likely cannot be retrieved.

The metadata to be embedded is converted into a spatial configuration of points (typically about 10). A 2-D PRN pattern is then generated from a (secret) key or seed. The PRN pattern is usually twice the size of the image to be tagged. To embed 64 bits of metadata with 20 bits of checksum, we embed 10 PRN patterns and use positional encoding as described in US patent 7158653 [37]. We note that other encoding methods are possible. According to Cox [38] spread spectrum watermarks with N bit binary string occupy a key-space of about $2^{0.12n}$. In the case of 2-D spread spectrum patterns $n \approx 256^2$ resulting in an effective key size of about 8k bits; cryptographic by most standards. The 10 copies of the PRN are then added together with the position offsets defined by configuration of points. The template and the message patterns are added together. We choose to have equal weights for the template and message patterns, although certain applications may favor weaker template patterns. Each LHF and PRN is set to have an RMS of 1. The template RMS is then approximately 2, and the total embedding RMS approximately √14.

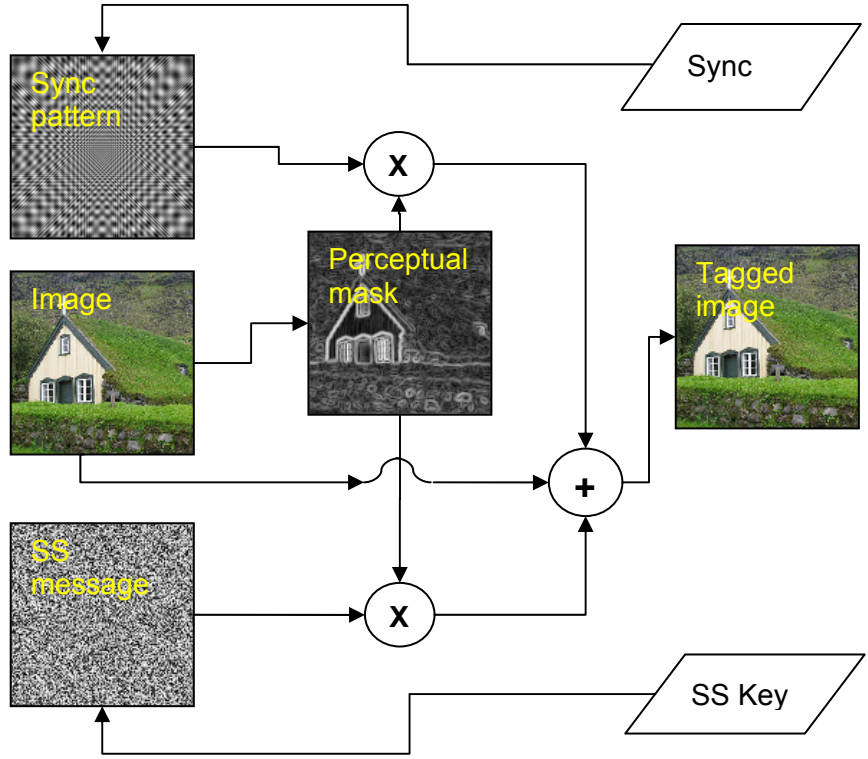

Fig.7　　Tag embedding process

## 6.2　Perceptual Masking

Now the composite pattern is multiplied by the perceptual mask. After much investigation we arrived at a pixel-wise perceptual mask based on a measure of the local image structure:

$$p(x_m, y_n) = \beta \left(1 + \sum \alpha_{k,l} \left|\nabla g_{m-k, n-l}\right|\right) \qquad (32)$$

We use a 9x9 Hann-weighted window defined by the coefficients $\alpha_{k,l}$. The discrete gradient is here defined by adjacent pixel difference in x and y. Watermark embedding is defined as pixel-wise addition

$$g(x_m, y_n) + p(x_m, y_n).w(x_m, y_n) \qquad (33)$$

The perceptual masking is, coincidentally, equivalent to embedding at a near constant Structural Similarity (SSIM) quality level [39-41]. The coincidence is, perhaps, less surprising considering the close mathematic connection with gradient-based image quality perception [39]. For the current implementation the coefficient $\beta$ is chosen to give a minimum composite signal with RMS of 1 gray-level in image regions with little structure (such as blue skies and smooth skin). The coefficients $\alpha_{k,l}$ are also chosen to give composite RMS values of 10 gray-levels in highly complex image regions (such as high contrast textures). In this instance the watermark is added to the luminance of a 24 bit RGB image.

## 6.3　Template Detection

An image for tag detection is shown at the top left of the flowchart in Fig.8. The first operation is to down-sample by factors of two in both directions until the diagonal is in a range between 1024 and 512 square image diagonals. Processing for large images is thereby reduced, with little effect on the embedded signal (because the embedding is optimized for this size range by simply excluding higher frequency components).
An inverse perceptual mask is estimated from the raw image data. The raw image and the inverse mask are multiplied to produce an enhanced image. This image is then 2-D Fourier transformed and then pseudo polar transformed. The pseudo-polar transform is multiplied by the conjugate MM characterized by an imaginary index (or sync key). The 2-D product image is then 1-D Fourier transformed (along the radial direction) to reveal in the magnitude a correlation peak map, similar to that in Fig.5. Inverse perceptual masking is found to significantly increase signal detection sensitivity. Further improvement results from local Fourier phase enhancement [29].

The N (typically 20) most likely peaks are identified. Sub-pixel shifts of peaks give reduced peak heights, so we use a local zooming operation (based on FFT up-sampling] in the immediate vicinity of each complex peak to regain full signal magnitude [42], further improving detection sensitivity. All permutations are tested against the known quartet of ratios using the LSF criterion, and the best fitting permutation yields a six parameter affine transformation.

### 6.4    Message Decoding

Message decoding uses standard watermarking techniques. The detector PRN pattern is generated from the input key or seed. To detect a distorted PRN we use the computed affine parameters from the template detection to pre-distort the matched filter. It is more efficient to generate the PRN as a phase only complex function in the Fourier domain and then apply the dual affine distortion in the Fourier domain. Correlation detection is implemented by 2-D multiplication of the FFT input image (after inverse perceptual masking), followed by an inverse FFT. The result is an array of sharp peaks. The configuration of peaks is tested for consistency and reliability and then decoded to reveal the tagged message. Typically we use 3 PRNs to define a coordinate system, 1 PRN to encode the number of expected message peaks (6 in this instance), and 6 PRNs to encode the actual (80 bit) message. Larger messages use more PRNs.

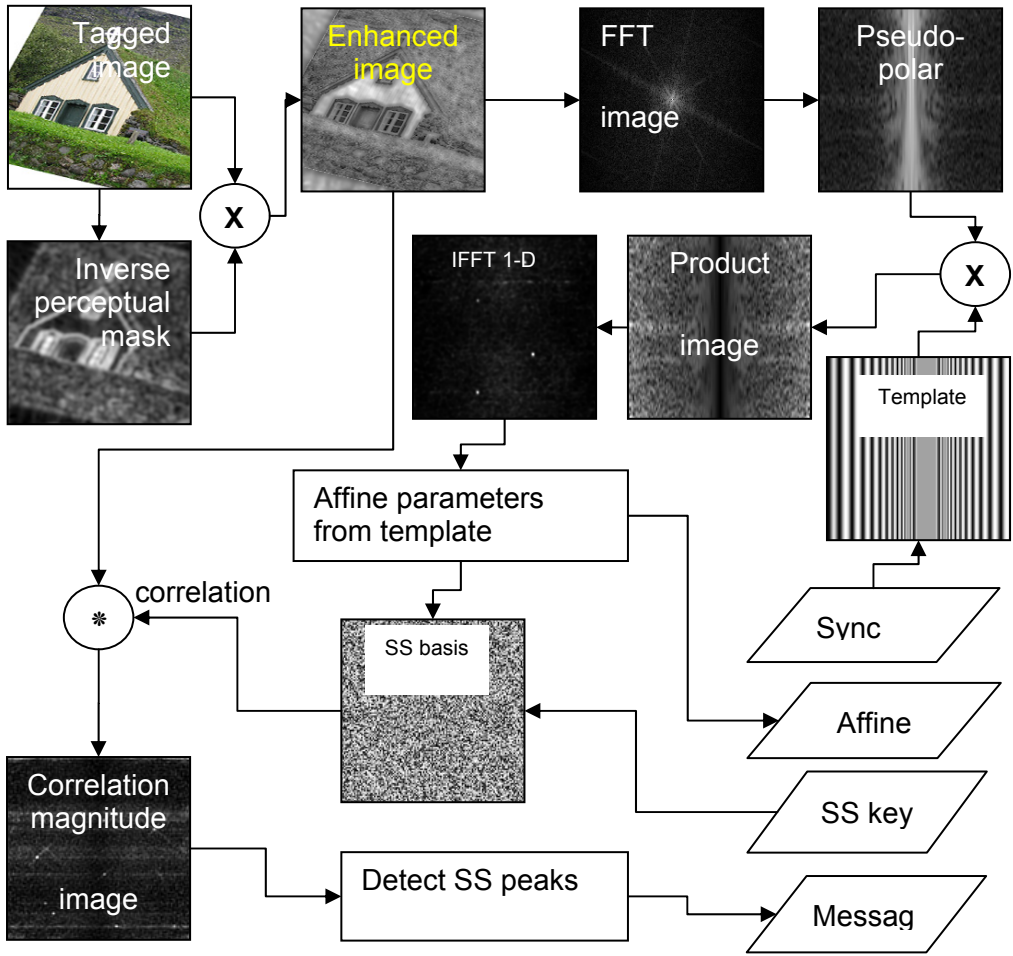

Fig.8    The tag detection process

## 7    System Performance

Stirmark [43, 44] is the de facto standard for characterizing WM robustness. Affine invariance can be attacked by non-affine geometric distortions and image row/column removals. Petitcolas and Anderson [45] suggest that a Peak Signal to Noise Ratio (PSNR) of 38 dB or greater corresponds to an imperceptible signal. They also note that PSNR is a very poor predictor of noise perceptibility. More recently Wang and Bovik [46] have proposed a perceptual image quality measure called Structural Similarity or SSIM that more closely correlates with subjective image quality and is relatively easy to compute. It is also possible to embed watermarks at a fixed fidelity level as described by Li and Cox [3, 47]. The nominal setting of our mask corresponds to a mean SSIM (MSSIM) over each image of $\geq 0.99$; judged to be barely perceptible. PSNR is then in the range 34dB to 45dB. PSNR 45dB corresponds to images dominated by smooth color gradations and little texture. PSNR 34dB corresponds to highly textured images.

Robust, affine distortion resistant watermarking schemes based on autocorrelation have been implemented by Philips and Kodak [48, 49]. How does our method compare? The main difference is that tiled PRN watermarks have obvious Fourier signatures; distinct FFT magnitude peaks in a regular grid. These peaks are easily detected and replaced by surrounding magnitude values. A corresponding attack on our WM is conservatively estimated to take four orders of magnitude more computation, because of the various keys that must be searched.

### 7.1 Stirmark results

Watermarks were embedded in images from a library. Owing to unresolved copyright issues our data set is not a public image library. We have tried to make the test representative, and our images broadly match the selection of Petitcolas [50]. Images are in the 1 to 3 megapixel range with just one image of 0.25 megapixels (a color image of Lena). Fig.9 shows the library of 48 images.

**Table 1. Stirmark results on 48 test images. Over 5000 distorted images were tested.**

| Stirmark 4 Benchmark Distortion | Pass rate (%) | Comments |
|---|---|---|
| Noise 20 (PSNR 16dB, MSSIM ~0.38) | 47 | Bad image quality at this noise level |
| Affine (1,2,3,4,6,7,8) | 100 | |
| Affine (5) | 98 | Failure due to edge/image structure |
| Convolution Filter 1* | 98 | Filter causes severe saturation |
| Convolution Filter 2* | 92 | Filter produces almost black images |
| Cropping 75x75% | 100 | 44 % of image removed |
| Cropping 50x50% | 98 | 75% of image removed |
| Median Cut 3 | 100 | |
| Median Cut 5 | 92 | Visible detail loss |
| Median Cut 7 | 33 | Substantial image detail loss |
| Median Cut 9 | 17 | Severe image detail loss |
| JPEG 15 | 73 | Very poor quality image |
| JPEG 20 | 81 | Very poor quality image |
| JPEG (25,30,35,40,50,100) | 98 | One small image fails (512x512) |
| Rescale (50,75,90,110,150,200 %) | 100 | |
| Rotate (-0.25,-1,-2,2,5,90°) | 100 | |
| Rotate (-0.5,0.75,1,10,15°) | 98 | Failure due image structure clash |
| Rotate (30°) | 96 | Failure due to edge/image structure |
| Rotate (45°) | 94 | Failure due to edge/image structure |
| Rotate/Crop (-0.25,-0.5,-1,-2°) | 100 | |
| Rotate/Crop (0.25,0.5,1,2°) | 100 | |
| Rotate/Scale (-0.25,-0.5,-1,-2) | 100 | |
| Rotate/Scale (0.25,0.5,1,2) | 100 | |
| Remove lines (10,20,30,40,50,60,70,80,90,100) | 100 | |
| Small random distortion (0.95,1,1.05,1.1) | 2* | *Stirmark software zeroed all images except the first, so this test is invalid and should really be discounted. |

We found larger images caused memory errors in the Stirmark processing, so we used down-sampled compact camera images. Table 1 shows a summary of results. For the 48 image database inter-image MSSIM variation is in the range 0.988 -0.9996, whilst the intra-image SSIM variation is in the range 0.895 – 1.0000.

The method performs well in terms of the *robustness-capacity-imperceptibility* watermark triangle, shown in Fig. 10. In two corners (*robustness-imperceptibility*) we are probably near the extremes of performance. In the third corner (*capacity*) we have $2^{64}$ unique identifiers ($2^{128}$ sufficient for a Universal Unique Identifier (UUID)). Typical embedding and detection times for a one megapixel image are 2.3 and 3.7 seconds respectively on one core of a 3 GHz PC. Four megapixel times are 5.8s and 4.0s.

The only Stirmark attack (of those which maintain acceptable image quality) our method completely fails is the random small geometric distortion. Although in this instance it was due to a Stirmark software problem, we acknowledge that our method is intrinsically susceptible to non-affine warps. Any non-affine warp which changes the local image magnification by more than 0.1% from the mean will reduce the correlation peaks in a one megapixel image by at least 50% and thus halving detectability.

The Stirmark noise attacks at ≤16dB PSNR actually makes the watermarked images unusable, and so, we suggest, can also be discounted. The tags survive most Stirmark attacks that also maintain reasonable perceived image quality. The main exception is random (non-affine) warping which our watermarks are not explicitly designed to resist. A few failures seem to be related to clashes between the alignment patterns and the image structure or the zero padding of the image boundary region after geometrical distortion. It is possible to prevent such failures in either the embedding or detection stages, but we have not implemented such improvements in the current version of our software.

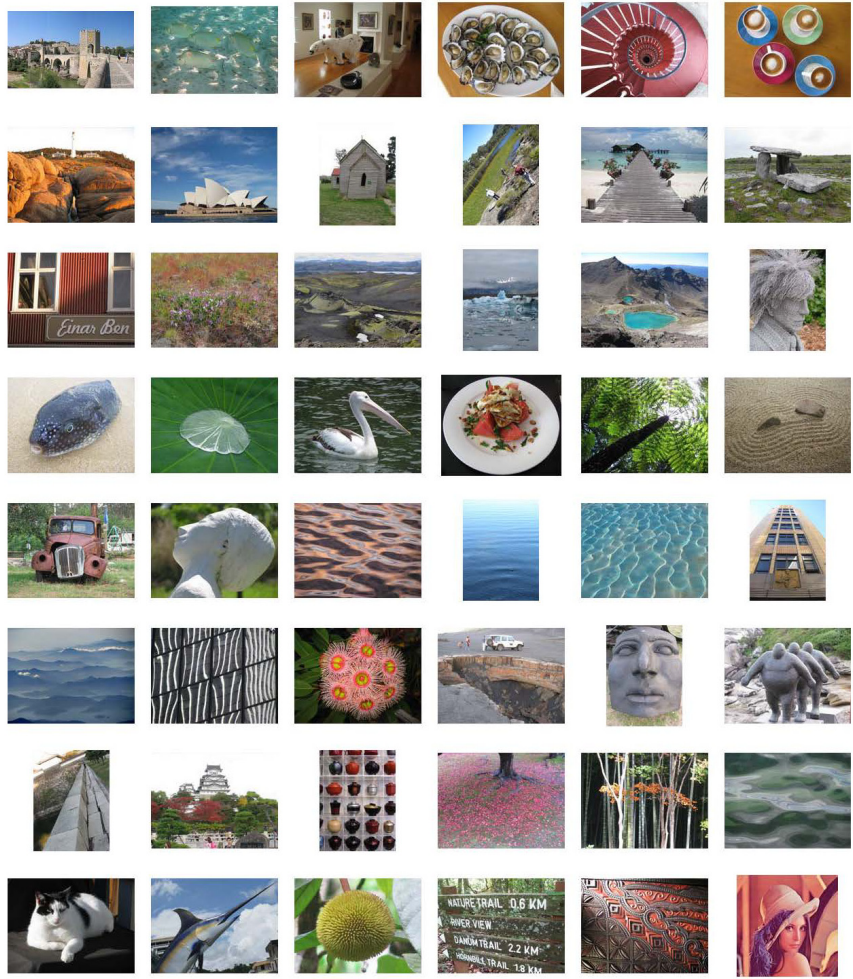

Fig.9    Private library of 48 images for Stirmark testing.

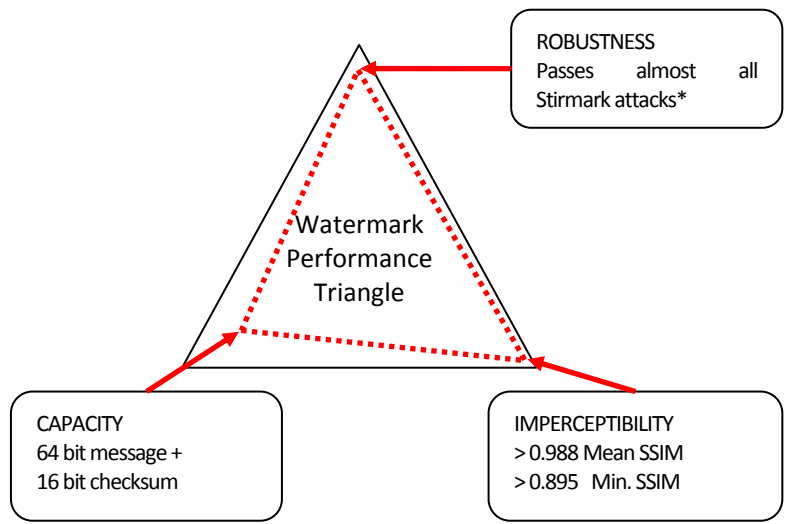

Fig.10    The robustness/capacity/imperceptibility performance triangle.

Two animations illustrating the robustness of the template detection process have been made available with the online version of this paper. Fig 11a (Media1) shows an upper test image, undergoing various distortions, with the detected peaks shown on the lower side. Fig 12b (Media 2) shows the same distortion sequence, but the image gamma has been changed to enhance the low-level background clutter in the peak detection process. The template is embedded at the sub-perceptible level of 44dB PSNR, with MSSIM of 0.99996 and a minimum SSIM of 0.99929. The animated distortion sequence is as follows:

- X and Y aspect ratio changes
- Color saturation from 0.0 to 2.0,
- Gamma change from 0.3 to 2.0,
- JPEG quality factor from 100 to 5,
- Additive uniform intensity noise from 0 to 50% peak ,
- Rotation +/- 5 degrees,
- Scaling from 0.60 to 1.64.

The peaks become difficult to distinguish only at the extreme range of gamma, jpeg quality, and additive noise.

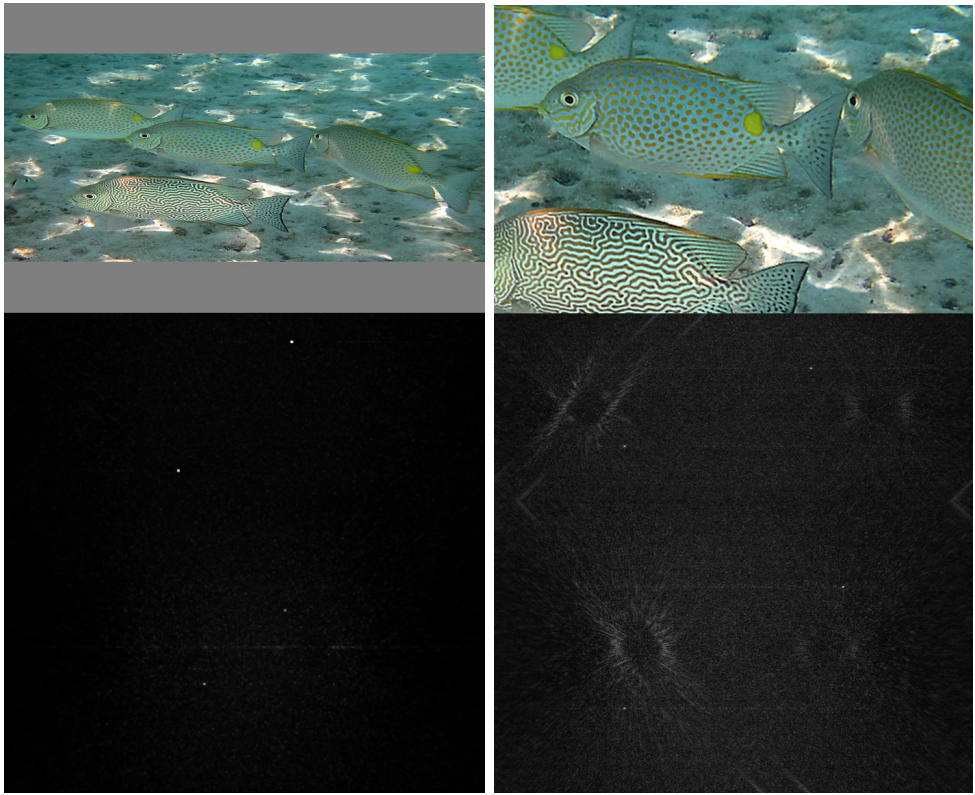

(a)  (b)

Fig.11a. Sequence (Media 1) of Stirmark-like image distortions (upper) and corresponding four template correlation peaks (lower) shown in linear intensity scale. (Peaks have been morphologically dilated to compensate optical dot gain in the print version of this figure.)

Fig.11b. Sequence (Media 2) of Stirmark-like image distortions (upper) and corresponding template correlation peaks (lower) shown with intensity gamma = 0.5 to emphasize the underlying clutter.

## 8    Discussion

We have described a new method for embedding imperceptible tags in digital images. Novelty resides in the invariant template design and detection. The Stirmark results are encouraging for high quality images in the megapixel size range. Printing and scanning performance were paramount in early development and it was found that the tag survives when the overall print-scan process yields visually acceptable quality images (but these results are not presented here). Overall our tagging method compares favorably against most published blind-watermarking techniques and only fails attacks which excessively compromise image quality or attacks that might be considered malicious or non-accidental.

We have taken advantage of a wide selection of mathematical invariances in our work: the shift invariance of correlations and Fourier transforms, the rotation invariances of the Radon transform and the projection-slice theorem, the affine invariance of line intersection ratios, the dilation invariance, self-transform, and orthogonal correlation properties of homogeneous functions. Computational advantage is taken of the method's intrinsic scale invariance by down-sampling all images to a convenient size before applying the tag generation, detection and decoding process. There are a number of possible extensions to this tagging method, in particular video. The method can easily avoid collusion attacks by systematically changing rotation, scale, translation and the phase of the template pattern for every video frame. Note that preliminary results from this research were presented in 2009 [51].

**Conflict of Interest**

The authors are not aware of any conflict of interest connected with the publication of this paper. The research was conducted during their normal employment by Canon Information Systems Research Australia Pty Ltd. Several of the novel and inventive aspects of the work presented are covered by patents, and for completeness these are cited in the text. The watermark benchmarking software Stirmark was used with permission of the owners, and the relevant publications cited accordingly.

**Appendix A1    Spatial interpretation of O'Ruanaidh/Pun RST algorithm**

The O'Ruanaidh/Pun RST invariant watermark [9] is normally defined as a pseudorandom 2-D signal embedded in a Fourier –Mellin (FM) domain. By working backward from the FM domain we can determine the equivalent spatial domain signal. We consider a single point in the FM domain represented by a Dirac delta function. Recalling that the sequence for the improved O'Ruanaidh algorithm is:
- Spatial domain image
- Fourier transform
- Take magnitude, keep old phase
- Log-polar coordinate transform (LPM)
- Fourier transform
- Add 2-D PRN (this is the FM domain): a single weighted delta $\varepsilon_n \delta(\sigma - \sigma_n, \gamma - \gamma_n)$
- Inverse Fourier transform: a single weighted exponential $\varepsilon_n \exp[+2\pi i(\sigma_n \rho + \gamma_n \varphi)]$
- Inverse LPM: yielding a single weighted LRHF centered on the DC frequency $\varepsilon_n q^{2\pi i \sigma_n} \cdot \exp[2\pi i \gamma_n \varphi]$
- Reconstruct from new magnitude and old phase: LRHF multiplied by original image Fourier phase $\exp[2\pi i \psi] \cdot \{\varepsilon_n q^{2\pi i \sigma_n} \cdot \exp[2\pi i \gamma_n \varphi]\}$
- Inverse Fourier transform gives the convolution of a phase-only [52] (i.e. edge enhanced) image and a spatial LRHF.

Hence each delta in the FM domain maps to a perceptually weighted (i.e. edge enhanced) image blurred by an oscillating LRHF. Deltas at other locations map to LRHFs with different radial and azimuthal indices, but the same centers (see [14, 51] for examples of typical LRHFs).

So, in summary, the RST watermark consists of a phase-only image convolved with a pseudorandom sum of centered LRHFs. The phase-only image serves as a perceptual mask for the embedded LRHFs and explains the importance of retaining the original Fourier phase in this scheme. As far as we are aware, the spatial domain interpretation of the O'Ruanaidh algorithm has not been published before, hence its inclusion here. The end result is distinctly unlike our spatially embedded signal.

**Appendix A2    Spatial interpretation of Lin/Wu/Bloom/Miller/Cox/Lui RST algorithm**

The Lin/Wu/Bloom/Miller/Cox/Lui RST invariant watermark [8] is, like that of O'Ruanaidh/Pun, defined in a domain related to the Fourier and log-polar transforms. The basic sequence in this instance is as follows:
- Spatial domain image
- Fourier transform
- Log-polar coordinate transform (LPM)
- Sum logs of values in each column (each column correspond to different polar angle)
- Add signal by distributing uniformly over column, and thus changing column log-sum
- Make sure signal is Fourier Hermitian, hence real spatial.
- Optimize embedding by an iterative loop of correlation detection and incrementing a target vector (note that the vector embedded is typically a pseudorandom vector).

The spatial domain interpretation would appear to be quite impenetrable. However a few observations help clarify the process. Firstly the projection means that the basis functions in the FT-LPM (i.e. FM) domain have angular variation only. This automatically makes them scale invariant. Note that a full analysis requires half a dozen pages, but only the outline is given here. Essentially the signal generated in the FM domain has angular variation only and is added to the log Fourier magnitude. The signal is constrained to Hermitian symmetry. Like the O'Ruanaidh/Pun scheme, the original image Fourier phase is maintained. The end result is an embedded signal which has annular support in the Fourier domain and exhibits pseudorandom variation in the azimuth and no variation radially. The signal is additive in the log-magnitude domain which makes it multiplicative in the magnitude domain. In the spatial domain this corresponds to a constant radial, pseudorandom azimuthal function *convolved* with the original image. This can be interpreted as a scale invariant function (with unique and detectable pseudorandom angular variation) perceptually weighted by image convolution. Again, as far as we are aware, this spatial domain interpretation has not been published before, hence its inclusion here to clearly differentiate from our direct spatial embedding method.